# Component SPD Matrices: A lower-dimensional discriminative data descriptor for image set classification

Kai-Xuan Chen

School of IOT Engineering, Jiangnan University
Wuxi, China
kaixuan_chen_jnu@163.com

Xiao-Jun Wu

School of IOT Engineering, Jiangnan University
Wuxi, China
xiaojun_wu_jnu@163.com

*Abstract*—In the domain of pattern recognition, using the SPD (Symmetric Positive Definite) matrices to represent data and taking the metrics of resulting Riemannian manifold into account have been widely used for the task of image set classification. In this paper, we propose a new data representation framework for image sets named CSPD (Component Symmetric Positive Definite). Firstly, we obtain sub-image sets by dividing the image set into square blocks with the same size, and use traditional SPD model to describe them. Then, we use the results of the Riemannian kernel on SPD matrices as similarities of corresponding sub-image sets. Finally, the CSPD matrix appears in the form of the kernel matrix for all the sub-image sets, and $CSPD_{i,j}$ denotes the similarity between $i$-th sub-image set and $j$-th sub-image set. Here, the Riemannian kernel is shown to satisfy the Mercer's theorem, so our proposed CSPD matrix is symmetric and positive definite and also lies on a Riemannian manifold. On three benchmark datasets, experimental results show that CSPD is a lower-dimensional and more discriminative data descriptor for the task of image set classification.

## I. Introduction

The image set classification, the task of classification based on image sets, has received wide attentions in the domain of artificial intelligence and pattern recognition [1],[2],[3],[4],[5],[6],[7],[8]. The image set contains a large number of images that are under different environments, so it can offer more robust and discriminative features than the single-shot image[9],[10],[11],[12],[13]. For the task of image set classification, representations of image sets commonly include, Gaussian mixture model [14], linear subspaces [1], covariance descriptors(CovDs)[7],[8], among which, CovDs have been widely applied in virus recognition [13], object detection [4], gesture classification [5]. The traditional SPD model is arising in the form of CovDs, which lie on a non-linear manifold known as SPD manifold.

The dimensionality of traditional SPD matrices arising from covariance descriptors [2],[4],[5],[6],[7],[8] used for image set classification is relatively high. Although high-dimensional data descriptors provide enough information, the curse of higher-dimensional data causes a lot of computations and has a poor effect on efficiency of the algorithms. The DR (dimensionality reduction) is always imperative in computer vision and machine learning. The classical methods, such as PCA(Principal Component Analysis)[16] and LDA(Linear Discriminant Analysis)[17] are pervasive in various applications. Because the SPD matrices lie on a non-linear manifold, these methods used in Euclidean space are not suitable for analyzing the SPD matrices. Recently, considering the Riemannian structure of SPD matrices, the work of DR has been extended to the space spanned by SPD matrices. The BCM (Bidirectional Covariance Matrices) [8]and SPDML[2] are the DR methods on the SPD manifold. BCM is a two-directional two-dimensional PCA[18] method directly working on the SPD matrices to obtain low-dimensional descriptors. SPDML[2] embed the high-dimension SPD matrices into a lower-dimensional and more discriminative SPD manifold through a projection matrix.

In this paper, we propose a new framework to obtain low-dimensional and more discriminative descriptors for representing image set. Now let $S$ be an image set and we assume there are $n$ samples in the set, $S = [s_1, s_2, \ldots s_n]$, where $s_i \in R^D$ represents the $i$-th image in the set. the traditional SPD model, which arises in the form of Covariance descriptors[2],[4],[5],[7],[8], will give a $D \times D$ SPD matrix for the representation of image set. $SPD_{i,j}$ denotes the covariance between $i\text{-}th$ dimension and $j\text{-}th$ dimension of the all images in image set, a.k.a. $i\text{-}th$ row and $j\text{-}th$ row of the image set matrix $S$. In our minds, we want to represent image set by describing the similarity between regions of the image set instead of the dimensions. In our CSPD model, we firstly divide the image set into $d \times d$ blocks, each block is a square with the same size. For the sub-image sets, we use covariance descriptor to represent the data in the sub-image sets.

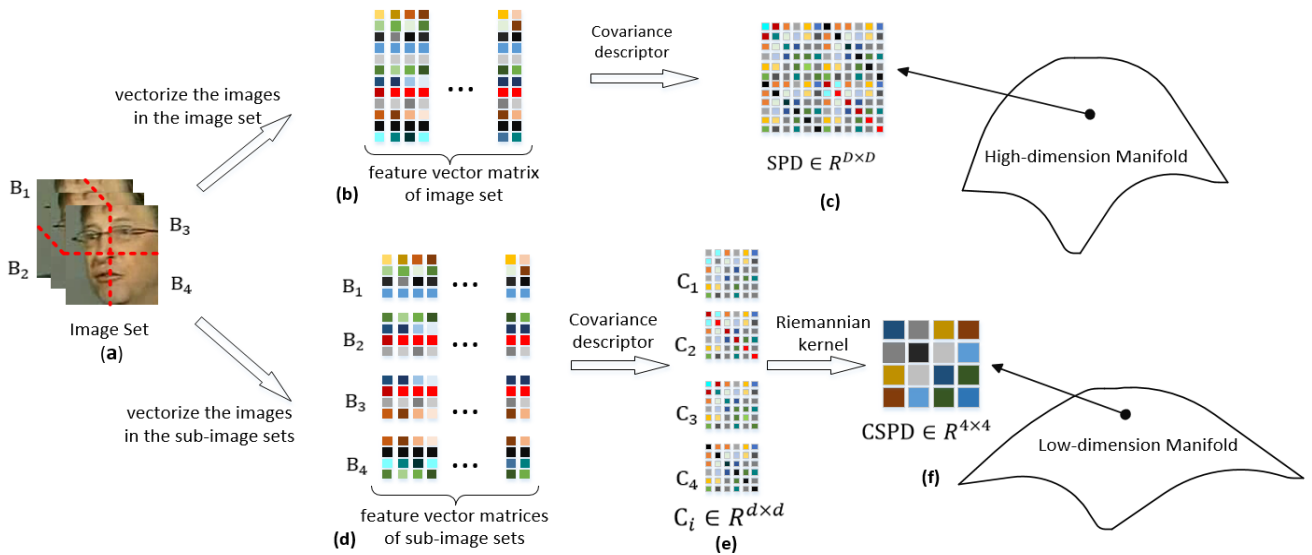

Fig.1. The flow chart of the traditional SPD model and our CSPD model. For an image set, the traditional SPD model follows the way (a)-(b)-(c), and the resulting SPD matrix is computed by covariance descriptor and lies on a non-linear geometry structure named SPD manifold. Our approach Component Symmetric Positive Definite(CSPD) model follows the way (a)-(d)-(e)-(f) to firstly divide the image set into square blocks with the same size, and the representation of *i*-th sub-image set B*i* is obtained by traditional SPD model. Then, we use the result of Riemannian kernel to describe the similarity between the sub-image sets, and the final CSPD appears in the form of the Riemannian kernel matrix of the representations of sub-image sets.

$CSPD_{i,j}$ denotes the similarity between *i*-th sub-image set and *j*-th sub-image set. Due to the number of the blocks is $d^2$ and $CSPD_{i,j} = CSPD_{j,i}$, the form of CSPD descriptor is a symmetric matrix with the dimensionality being $d^2 \times d^2$. At last, in order to ensure the positive definiteness of this symmetric matrix, we use the result of the Riemannian kernel function on SPD matrices corresponding to the sub-image sets as the similarity between sub-image sets. Figure 1 shows the entire produce of our approach and traditional model for image set.

The remaining of this paper is organized as follows: In Section II, we give a brief overview on the geometry of SPD manifold and some classical related Riemannian metrics. In Section III, we present the model of original SPD and our proposed CSPD, and introduce some SPD manifold-based classification algorithms which are used in the experiments of this paper. In Section IV, we present the experimental results with the average accuracies and standard deviations. In our experiments, we conduct experiments on three tasks of object categorization, hand gesture recognition and virus cell classification. Moreover, the experiments show that our model CSPD has better recognition rates and improves the efficiency of the classification algorithms. In Section V, we present our conclusions and future directions.

## II. RELATED WORK

In this section, we will give an overview on the geometry of SPD manifold and some classical related Riemannian metrics. In this paper, we will take the following notation: $S_n^+$ is the space spanned by real $n \times n$ SPD matrices, $S_n$ is the tangent space spanned by real $n \times n$ symmetric matrices at the point of identity matrix $I_n \epsilon R^{n \times n}$. $T_P S_n^+$ is the tangent space spanned by real $n \times n$ symmetric matrices at the point of $P \epsilon S_n^+$.

### A. SPD manifold

The SPD matrices have been proved to be the powerful data representation approach for images or image sets via covariance[7],[8] or region covariance[15] descriptors. The space spanned by the SPD matrices does not satisfy the scalar multiplication axiom of the vector space. For example, the result matrix via multiplying an SPD matrix by a negative scalar does not lie on $S_n^+$[11]. The similarity between two SPD matrices computed by the Euclidean metrics is not reasonable and Riemannian metrics have been proven to get a better effect on the SPD matrices. As studied in[2], the SPD manifold spanned by SPD matrices is one kind of Riemannian manifolds and forms the interior of a convex cone in the Euclidean space.

A variety of Riemannian metrics of SPD manifold have been proposed. In particular, the AIRM (Affine Invariant Riemannian Metric)[2],[8] is the mostly studied descriptor which is the geodesic distance between two SPD matrices, and has the property of invariance to affine transformations. The Stein divergence and Jeffrey divergence[2],[10], which are efficient metrics akin to AIRM to measure geodesic distance between two SPD matrices, are Bregman divergence for some special seed function. The LEM (Log-Euclidean Metric)[7],[8] obtains the similarity between two SPD matrices through computing the distance in the space of matrix logarithm which is the tangent space at the point of identity matrix. Then, we will introduce AIRM and LEM in detail, and these two metrics will be used in our experiment.

### B. Affine Invariant Riemannian Metric

The $S_n^+$ can be viewed as a convex cone in the $n(n+1)/2$ dimensional Euclidean space[2]. The similarity between two SPD matrices on the manifold can be described by the length of geodesic curve, which is analogous to the straight line between two points in the vector space. The AIRM[2],[8] is one of the most popular Riemannian metrics on the SPD manifold, and it measures the similarity between two points on SPD manifold by computing the geodesic distance between them. For point $P$ on the SPD manifold, The AIRM can be defined through two tangent vectors $u, v \epsilon T_P S_n^+$:

$$<u,v>_P \triangleq <P^{-\frac{1}{2}}uP^{-\frac{1}{2}}, P^{-\frac{1}{2}}vP^{-\frac{1}{2}}> = tr(P^{-1}uP^{-1}v) \quad (1)$$

The geodesic distance $d_{AIMR}$ between two points $X$ and $Y$ on SPD manifold computed by AIRM can be written as:

$$d_{AIMR}(X,Y) = ||\log\left(X^{-\frac{1}{2}}YX^{-\frac{1}{2}}\right)||_F \quad (2)$$

where $||\cdot||_F$ denotes the Frobenius norm, $\log(\cdot)$ is the matrix logarithm operator.

*C. Log-Euclidean metric*

LEM(Log-Euclidean metric)[4],[7],[8],[11] is a bi-variant Riemannian metric coming from the Lie group multiplication on SPD matrices[11]:

$$\odot: S_n^+ \times S_n^+ \to S_n^+$$
$$X \odot Y = exp(\log(X) + \log(Y)) \quad (3)$$

where $X$ and $Y$ lie on SPD manifold. The distance $d_{LogED}$ between these two SPD matrices computed by this metric can be written as:

$$d_{LogED}(X,Y) = ||\log(X) - \log(Y)||_F \quad (4)$$

where $\log(\cdot)$ is the matrix logarithm operator, $||\cdot||_F$ denotes the Frobenius norm. The results of LEM can be viewed as the distance of the points in the tangent space $S_n$ projected from SPD manifold $S_n^+$ by logarithm mapping[7],[8]:

$$\varphi_{\log}: S_n^+ \to S_n, Sp \to \log(Sp), Sp \epsilon S_n^+ \quad (5)$$

where $S_n$ is a vector space, and Figure 2 gives the conceptual illustration of logarithm mapping vividly. In compliance with Riemannian multiplication $\odot$ operater on SPD matrices, the scalar multiplication can be defined[11] as:

$$\lambda \otimes Sp = exp(\lambda \log(Sp)) = Sp^\lambda, Sp \epsilon S_n^+ \quad (6)$$

where $\lambda$ is a real scalar. $S_n^+$ is a vector space when endowed with the Riemannian multiplication $\odot$ and Riemannian scalar multiplication $\otimes$ [11]. Furthermore, the Riemannian kernel function can be represented by the Log-Euclidean inner product[7],[11]:

$$k_{LogE}(X,Y) = <X,Y>_{LogE} = tr(\log(X)\log(Y)) \quad (7)$$

For the all points $Sp_1, \dots, Sp_N \epsilon S_n^+$, $k_{LogE}$ is a symmetric function because of $k_{LogE}(Sp_i, Sp_j) = k_{LogE}(Sp_j, Sp_i)$. According to the paper [7], we have:

$$\sum_{i,j} a_i a_j k_{LogE}(Sp_i, Sp_j) = \sum_{i,j} a_i a_j tr(\log(Sp_i)\log(Sp_j))$$
$$= tr[\sum_{i,j} a_i a_j \log(Sp_i)\log(Sp_j)] = ||\sum_i a_i \log(Sp_i)||_F^2 \geq 0$$
$$a_i \epsilon R, \forall i \epsilon N \quad (8)$$

The Eq(8) gives the proof that Log-Euclidean kernel guarantee the positive definite property of the Riemannian kernel and satisfies the Mercer's theorem. The kernel matrix of the all points on SPD manifold is also a SPD(Symmetric Positive Definite) matrix.

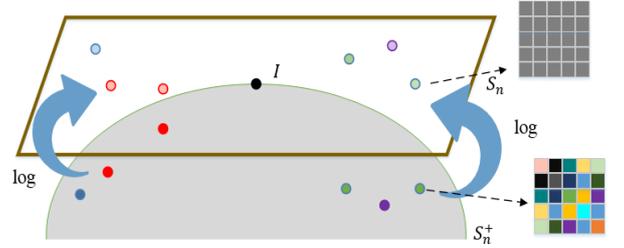

Fig 2. Logarithm mapping

III. COMPONENT SYMMETRIC POSITIVE DEFINITE MATRIX

In this section, we will recall the Original SPD model for image set obtained by covariance descriptors[7],[8] and introduce our model CSPD(Component Symmetric Positive Definite) model in detail.

*A. Original Symmetric Positive Definite model*

For an image set with $n$ images: $S = [s_1, s_2 \dots, s_n]$, where $s_i \epsilon R^D$ represents the $i$-th image sample of $D$-dimensional vector. Here, the covariance matrix[2],[7],[8] computed from the raw intensity of sample in the image sets:

$$C = \frac{1}{n}\sum_{i=1}^n (s_i - \bar{s})(s_i - \bar{s})^T = \frac{1}{n}SJ_nS^T \quad (9)$$

where $\bar{s} = \frac{1}{n}\sum_{i=1}^n s_i$ is the mean of all images which are represented by the $D$-dimensional vectors in the set $S$. $J_n = I_n - \frac{1}{n}1_n 1_n^T$ is the centering matrix and $1_n$ is a column vector of $n$ ones [2]. Also $J_n$ is a symmetric matrix, with the rank$(J_n) = n - 1$ and $J_n^2 = J_n$. In general, the number of the images in the set is often smaller than the dimensionality of the feature vector, thus a covariance matrix is not positive definite, so we need to add a small perturbation[7]:

$$C^* = C + \lambda \times tr(C) \times I \quad (10)$$

where $\lambda$ was set as $10^{-3}$ and $I$ is the identity matrix[7]. Now, the image sets are modeled as the SPD matrices which form the SPD manifold. In general, the dimensionality of the covariance descriptors is high, our model overcomes this limitation.

*B. Component Symmetric Positive Definite model*

For our model CSPD, we firstly divide image set into $d \times d$ square blocks with the same size. One block of image set is described by covariance descriptor(Eq.9), there are $d^2$ SPD matrices for all blocks in the image set. Our model is proposed to describe the relationship between the blocks of the image set.

For example, from the Fig.1 path of arrow (a)-(c)-(d), the image set was divided into $2 \times 2$ square blocks and form 4 sub-image sets:$B_1, B_2, B_3$ and $B_4$ firstly. Correspondingly, there are 4 covariance descriptors $C_1, C_2, C_3$ and $C_4$ for 4 sub-image sets. To this end, $CSPD \epsilon R^{4 \times 4}$ is a matrix describing the similarity between 4 sub-image sets. Here in Fig.1(c), $C_i$ lies on a higher-dimensional SPD manifold even though the dimensionality of the images in the blocks is lowers. In order to measure similarity between sub-image sets, we use the Log-Euclidean inner product(Eq.7) to represent the similarity of the covariance descriptors:

$$CSPD_{i,j} = k_{LogE}(C_i, C_j) = tr(\log(C_i)\log(C_j)) \quad (11)$$

Here, $CSPD_{i,j}$ means the similarity between $i$-th sub-image set and $j$-th sub-image set, and $CSPD_{i,j} = CSPD_{j,i}$. We use the Log-Euclidean inner product[7],[11] to represent the covariance descriptors, because it can ensure the positive definite property of the CSPD. The final CSPD appears in the form of the Riemannian kernel matrix of the covariance descriptors $C_i$ for all sub-image sets.

*C. Classification algorithms based on SPD manifold*

The NN (nearest neighbor) algorithm is one of the simplest methods for classification and regression in the domain of computer vision and pattern recognition. This classification algorithm classifies the input point to the class of the closest neighbor point and will show different accuracies under different geometric metrics. According to the literature [8], the NN classification algorithms based on AIRM and LEM are utilized to the SPD manifold, and these simple classification algorithms can clearly show the benefits of our CSPD model.

In the literature [7], the CDL(covariance discriminative learning) was proposed for image set classification. In this paper, the geometric properties of the Riemannian manifold are fully considered, and the classical classification algorithms are not directly utilized to the SPD manifold. It derives a kernel function that maps the SPD matrices from the Riemannian manifold to the Euclidean space through the LEM metric. With this mapping, the classical classification algorithms applied in the linear space can be exploited in the kernel formulation. LDA (linear discriminant analysis) and PLS (partial least squares) devoted to the linear space are considered in the literature [7] for the task of classification.

Lastly, we introduce the Riemannian sparse coding algorithm LogEKSR[11] which applies the sparse representation and dictionary learning to SPD matrices through mapping the SPD matrices into RKHS (Reproducing Kernel Hilbert Space) to obtain the sparse coefficients through Log-Euclidean kernels. Note that the Log-Euclidean kernels in this algorithm are the derivatives of Eq.7.

Log-E poly.kernel $\quad k_{p_n}(S, T) = p_n(<S, T>_{LogE}) \quad (12)$

Log-E exp.kernel $\quad k_{e_n}(S, T) = exp(p_n(<S, T>_{LogE})) \quad (13)$

Log-E Gaus.kernel $\quad k_g(S, T) = exp(-||S \odot T||^2_{LogE,A}) \quad (14)$

Where $p_n$ is a polynomial of degree n ≥ 1 with positive coefficients. For the Gaussian kernel, A is a diagonal matrix $A = diag\{\beta\}$ with $\beta > 0$, $k_g(S, T)$ reduces to a special form $exp(-||\log(S) - \log(T)||^2_F)$. According to the literature [11], these kernels are positive definite and meet the Mercer's theorem.

IV. EXPERIMENTAL RESULTS AND ANALYSIS

In order to verify the effectiveness of our model, we do experiments on the three tasks: object categorization, hand gesture recognition and virus cell classification. The three datasets are ETH-80[4], Cambridge hand gesture dataset(CG)[5], and Virus dataset[13] respectively. In our experiments, we compare the accuracies of our model CSPD with original SPD model under the same classification algorithms. Firstly, we take a most commonly used nearest neighbor classifier based on AIRM[2],[8] and LEM[4], [8],[11] which are introduced in section 2. The NN classifier is a simple method to display the advantage of our model. Secondly, we make use of classical Riemannian classification algorithms Log-E poly.kernel-based LogEKSR(Log-Euclidean Kernels for Sparse Representation)[11] and LDA-based CDL(Covariance Discriminative Learning)[7], which are the efficient methods on SPD manifold. Next, we give the naming of different algorithms:

- $NN - AIRM_{SPD}$ : AIRM-based Nearest Neighbor classifier for the SPD manifold spanned by original SPD matrices.
- $NN - AIRM_{CSPD}$ : AIRM-based Nearest Neighbor classifier for the CSPD manifold spanned by our proposed CSPD matrices.
- $NN - LogED_{SPD}$ : LEM-based Nearest Neighbor classifier for the SPD manifold spanned by original SPD matrices.
- $NN - LogED_{CSPD}$ : LEM-based Nearest Neighbor classifier for the CSPD manifold spanned by our proposed CSPD matrices.
- $CDL_{SPD}$: CDL-based classifier for the SPD manifold spanned by original SPD matrices.
- $CDL_{CSPD}$: CDL-based classifier for the CSPD manifold spanned by our proposed CSPD matrices.
- $LogEKSR_{SPD}$: LogEKSR-based classifier for the SPD manifold spanned by original SPD matrices.
- $LogEKSR_{CSPD}$ : LogEKSR-based classifier for the CSPD manifold spanned by our proposed CSPD matrices.

In our experiments, we re-size all the images into $24 \times 24$ and the image set can be divided into $2 \times 2, 3 \times 3, 4 \times 4, 6 \times 6, 8 \times 8$ and $12 \times 12$ blocks. For such image size setting, the dimensionality of the original SPD is $576 \times 576$. Instead, the dimensionality of the CSPD will be $4 \times 4, 9 \times 9, 16 \times 16, 36 \times 36, 64 \times 64$ and $144 \times 144$. Here, just from the view of dimensionality of the two kinds of data descriptors, our approach has lower-dimensional data representation. Next, we will use the results of the experiments to verify the discrimination of our model.

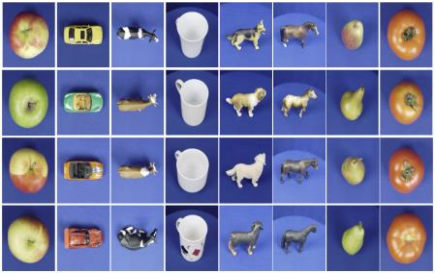 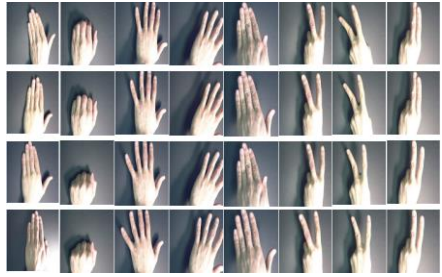 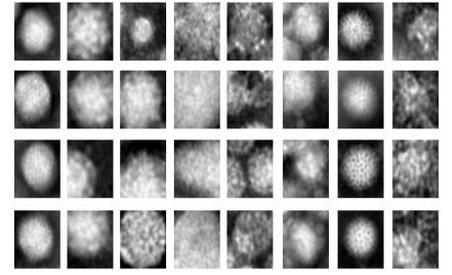

Fig.3. Images in ETH-80 dataset      Fig.4. Images in Cambridge gesture dataset      Fig.5. Images in Virus cell dataset

Table 1: Recognition rates and standard deviations for the ETH-80 dataset

| Method | Accuracy |
| --- | --- |
| $NN - AIRM_{SPD}$ | $58.22 \pm 6.35$ |
| $NN - AIRM_{CSPD}$ | $84.92 \pm 4.54$ |
| $NN - LogED_{SPD}$ | $64.48 \pm 6.25$ |
| $NN - LogED_{CSPD}$ | $87.52 \pm 3.85$ |
| $CDL_{SPD}$ | $78.66 \pm 7.07$ |
| $CDL_{CSPD}$ | $88.92 \pm 3.95$ |
| $LogEKSR_{SPD}$ | $86.94 \pm 4.58$ |
| $LogEKSR_{CSPD}$ | $\mathbf{89.92 \pm 3.84}$ |

Table 2: Recognition rates and standard deviations for the CG dataset

| Method | Accuracy |
| --- | --- |
| $NN - AIRM_{SPD}$ | $51.77 \pm 2.47$ |
| $NN - AIRM_{CSPD}$ | $76.39 \pm 1.81$ |
| $NN - LogED_{SPD}$ | $67.49 \pm 1.60$ |
| $NN - LogED_{CSPD}$ | $80.60 \pm 1.48$ |
| $CDL_{SPD}$ | $89.23 \pm 2.00$ |
| $CDL_{CSPD}$ | $90.84 \pm 1.20$ |
| $LogEKSR_{SPD}$ | $89.69 \pm 1.19$ |
| $LogEKSR_{CSPD}$ | $\mathbf{91.02 \pm 1.54}$ |

Table 3: Recognition rates and standard deviations for the Virus cell dataset

| Method | Accuracy |
| --- | --- |
| $NN - AIRM_{SPD}$ | $27.57 \pm 4.34$ |
| $NN - AIRM_{CSPD}$ | $33.67 \pm 6.33$ |
| $NN - LogED_{SPD}$ | $25.97 \pm 4.62$ |
| $NN - LogED_{CSPD}$ | $41.07 \pm 6.10$ |
| $CDL_{SPD}$ | $45.30 \pm 5.65$ |
| $CDL_{CSPD}$ | $\mathbf{54.50 \pm 7.38}$ |
| $LogEKSR_{SPD}$ | $47.13 \pm 4.58$ |
| $LogEKSR_{CSPD}$ | $53.77 \pm 6.44$ |

### A. Object Categorization on the dataset ETH-80

For the task of object categorization, we selected the ETH-80 dataset for experiments. The ETH-80 contains eight categories images of apples, pears, tomatoes, cows, dogs, horses, cups, and cars. Each class has 10 image sets, and each image set consists of 41 images from different angles. Fig.3. gives the part of images in the ETH-80 dataset. For each class, we randomly choose 2 image sets as training data, and the rest image sets were used as test data. We give the average accuracies and standard deviations of the 10 cross validation experiments.

Table 1 shows the performance of our model CSPD and original model SPD under the same classification algorithms. The results of our CSPD model with four different classification algorithms are on the premise of the image set being divided into $6 \times 6$ blocks. We can see that the results of NN classifier based on two introduced Riemannian metrics and CDL-based classifier are improved significantly by using our CSPD model. In particular, the these NN classification algorithms $NN - AIRM_{CSPD}$ and $NN - LogED_{CSPD}$ based on our model CSPD not only outperform $NN - AIRM_{SPD}$ and $NN - LogED_{SPD}$, but also outperform $CDL_{SPD}$ and $LogEKSR_{SPD}$ based on original SPD model. To the end, the accuracies of the $CDL_{CSPD}$ and $LogEKSR_{CSPD}$ are better than the $CDL_{SPD}$ and $LogEKSR_{SPD}$, and the $LogEKSR_{CSPD}$ achieves the best accuracy of 89.92% and the lowest standard deviation of 3.84.

### B. Hand Gesture Recognition

Cambridge hand gesture dataset composed of a set of high resolution color sequences acquired by the Senz3D sensor is an image sequence of hand gestures defined by 3 primitive hand shapes and 3 primitive motions. In this dataset, there are 900 image sets of 9 classes with 100 image sets in each class (see Fig.4 for example). For the task of hand gesture recognition, 20 image sets of each class were randomly selected as training data, and the rest image sets were chosen as test data. Ten-fold cross validation experiments were operated on this dataset.

We give the average accuracies and standard deviations of ten experiments in Table 2. The results of our CSPD model with four different classification algorithms are on the premise of the image set being divided into $6 \times 6$ blocks. For all the classifiers, the CSPD model has the advantages of higher recognition rates and lower standard deviations. Again we can see that the recognition rates of NN classification algorithms with CSPD model have obvious advantages over SPD model. In all the methods, $LogEKSR_{CSPD}$ achieves the best recognition rate of 91.02% and lower standard deviation of 1.54.

### C. Virus Cell Classification

The Virus dataset contains 15 categories, each category contains 5 image sets, each with 20 pictures taken from different angles. We arbitrarily choose 3 for training and the rest for testing. Figure 5 is the part of the images in Virus dataset, Table 3 shows the average recognition rates and standard deviations of ten experiments with respect to four algorithms on the Virus dataset.

Table 3 gives the results of the different methods with different image set descriptors SPD and CSPD. The results of our CSPD model with four different classification algorithms are on the premise of the image set being divided into $4 \times 4$ blocks. Note that the recognition rates of all methods with the CSPD model is higher than the SPD model. In particular, the accuracy of $NN - LogED_{CSPD}$ with CSPD model is similar to CDL-based $CDL_{SPD}$ with original SPD model. In the all methods, $CDL_{CSPD}$ achieves the best recognition rate of 54.50%.

### D. Effects of block number

Here, we will present the effects of block number on the average accuracies, standard deviations and running time under the same classification algorithm. Here, we do the experiment on the ETH-80 dataset as an example and give the next notations.

- $SPD^{OR}$: the data representation obtained by covariance descriptors
- $CSPD^{n \times n}$: the CSPD descriptor obtained by dividing the image into $n \times n$ blocks

*1) Effects of block number on average accuracies and standard deviations*

In order to display the effects of the block number, we will show the average accuracies of the 6 kinds of CSPD descriptors arising from the different segmentations of the image set and the original SPD obtained by covariance descriptor on ETH-80 dataset shown in Table.4, which shows the average accuracies of different data descriptors with the NN-AIRM (AIRM-based Nearest Neighbor classifier), NN-LogED (LEM-based Nearest Neighbor classifier), CDL and LogEKSR classification algorithms. The data in the row are the average recognition rates of the same data descriptor with the different classification algorithms, and the data in the column are the average recognition rates of the same classification algorithm with different data descriptors. From Table.4, we can see that the recognition rates of four classification algorithms with CSPD model are lower than the original SPD model when the image set was divided into $2 \times 2$ blocks. Finally, the four algorithms have a better recognition rates for all classification algorithms when the image set was divided into $6 \times 6$ square blocks.

Table 4 effects of block number on average accuracies

| Models | NN-AIRM | NN-LogED | CDL | LogEKSR |
|---|---|---|---|---|
| $SPD^{OR}$ | 58.22 | 64.48 | 78.66 | 86.94 |
| $CSPD^{2 \times 2}$ | 56.72 | 63.48 | 62.98 | 59.58 |
| $CSPD^{3 \times 3}$ | 73.73 | 79.89 | 81.02 | 81.63 |
| $CSPD^{4 \times 4}$ | 82.31 | 85.52 | 86.45 | 86.44 |
| $CSPD^{6 \times 6}$ | 84.92 | 87.52 | 88.92 | 89.92 |
| $CSPD^{8 \times 8}$ | 85.83 | 86.66 | 88.03 | 89.34 |
| $CSPD^{12 \times 12}$ | 86.86 | 84.19 | 87.52 | 88.69 |

In order to show the robustness of original SPD model and our proposed CSPD model on ETH-80 dataset, we give average standard deviations of ten experiments in Table.5. The data in the row are the standard deviations of the same data descriptor with the different classification algorithms, and the data in the column are the standard deviations of the same classification algorithm with different data descriptors. As shown in the Table.5, we can see that the standard deviations of our CSPD model are generally lower than the original SPD model with the same classification algorithm. Especially, the standard deviations of our CSPD model is lower than the original SPD model with the same classification algorithm when the image set was divided into $3 \times 3, 4 \times 4, 6 \times 6, 8 \times 8$ and $12 \times 12$ blocks.

Table 5 effects of block number on standard deviations

| Models | NN-AIRM | NN-LogED | CDL | LogEKSR |
|---|---|---|---|---|
| $SPD^{OR}$ | 6.35 | 6.25 | 7.07 | 4.58 |
| $CSPD^{2 \times 2}$ | 5.66 | 5.67 | 6.66 | 5.99 |
| $CSPD^{3 \times 3}$ | 5.47 | 4.18 | 4.40 | 4.37 |
| $CSPD^{4 \times 4}$ | 4.91 | 4.78 | 4.27 | 3.74 |
| $CSPD^{6 \times 6}$ | 4.54 | 3.85 | 3.95 | 3.84 |
| $CSPD^{8 \times 8}$ | 4.00 | 4.31 | 3.82 | 4.12 |
| $CSPD^{12 \times 12}$ | 4.24 | 4.23 | 4.18 | 3.72 |

According to the above two tables, we have the finding that our CSPD model used on ETH-80 dataset has higher recognition rates and lower standard deviations when the image set was divided into $6 \times 6$ blocks, and the results of classification algorithms with CSPD model in Table 1 are obtained on the premise of the image set being divided into $6 \times 6$ blocks. Similarly, we divide the image set of Cambridge hand gesture dataset into $6 \times 6$ blocks to obtain the results of CSPD model in Table 2. The results of CSPD model in Table 3 are obtained by dividing the image set of Virus cell dataset into $4 \times 4$ blocks. Here, we will not give the average accuracies and standard deviations under different data descriptors with the same classification algorithms for Cambridge hand gesture dataset and Virus cell dataset.

*2) Effects of block number on running time*

The dimensionality of our CSPD matrices is lower than original SPD matrices. This property is good for saving running time. We consider the efficiency of our CSPD model from two aspects: 1) the running time of different data representation models with the same classification. 2) the time of obtaining data descriptors from image set to SPD or CSPD.

Firstly, we give the Table 6 which shows the time needed from image set to data descriptors (SPD or CSPD). The unit of the time is second. As can be seen from Table 6, the time needed for CSPD is less than that of original SPD while image set being divided into $2 \times 2, 3 \times 3, 4 \times 4$ and $6 \times 6$ blocks. In general, the gap between the time needed for different descriptors is relatively small.

Table 6 time needed from image set to data descriptors

| Models | Time |
|---|---|
| $SPD^{OR}$ | 3.03 |
| $CSPD^{2\times2}$ | 2.80 |
| $CSPD^{3\times3}$ | 2.73 |
| $CSPD^{4\times4}$ | 2.47 |
| $CSPD^{6\times6}$ | 2.98 |
| $CSPD^{8\times8}$ | 3.17 |
| $CSPD^{12\times12}$ | 3.64 |

Secondly, we give the comparison of running time of different data descriptors with the classification algorithms in Table 7. The data in the row are the running time of the same data descriptor with the different classification algorithms, and the data in the column are the running time of the same classification algorithm with different data descriptors.

Table 7 Comparison time of different data descriptors with the same classification algorithm

| Models | NN-AIRM | NN-LogED | CDL | LogEKSR |
|---|---|---|---|---|
| $SPD^{OR}$ | 31063.0 | 4168.9 | 348.6 | 877.5 |
| $CSPD^{2\times2}$ | 13.6 | 3.4 | 8.1 | 2.6 |
| $CSPD^{3\times3}$ | 20.6 | 3.9 | 9.2 | 4.6 |
| $CSPD^{4\times4}$ | 37.9 | 5.5 | 10.0 | 5.9 |
| $CSPD^{6\times6}$ | 137.8 | 15.6 | 10.9 | 6.0 |
| $CSPD^{8\times8}$ | 422.0 | 42.2 | 13.0 | 12.9 |
| $CSPD^{12\times12}$ | 1634.1 | 211.6 | 33.4 | 58.0 |

As shown in the Table 7 where the unit of the time is millisecond, the advantages of our CSPD model are obvious with the same classification algorithm whatever the kinds of CSPD model. According to these two tables, we have the observation that the efficiency of classification algorithms with our approach of data descriptors has been greatly improved.

V. CONCLUSION

In this paper, we propose the CSPD(Component Symmetric Positive Definite) model to extract novel descriptors for image sets. The superior performance of our proposed CSPD is mainly demonstrated with more discriminative ability and lower dimensionality. For the property of discriminative ability, the recognition rates of CSPD are higher than that of traditional SPD while using the same classification algorithm. Especially, it can be expressed directly and effectively by the comparisons of the results from two Nearest Neighbor classification algorithms. For the property of lower dimensionality, the time complexity has been decreased and efficiency for algorithms have been improved significantly. For the future work, we will study more data descriptors for image set classification.